\documentclass[letterpaper]{article} % DO NOT CHANGE THIS
\usepackage{aaai2027}  % DO NOT CHANGE THIS
\usepackage[hyphens]{url}  % DO NOT CHANGE THIS
\usepackage{graphicx} % DO NOT CHANGE THIS
\usepackage{natbib}  % DO NOT CHANGE THIS AND DO NOT ADD ANY OPTIONS TO IT
\usepackage{caption} % DO NOT CHANGE THIS AND DO NOT ADD ANY OPTIONS TO IT
\usepackage{algorithm}
\usepackage{url}
\usepackage{enumitem}
\usepackage{algpseudocode}
\usepackage{adjustbox}
\usepackage{makecell}
\usepackage{multirow}
\usepackage{subcaption}
\usepackage{graphicx}
\usepackage{amsmath}
\usepackage{amssymb}
\usepackage{scrextend}
\usepackage{bm}

\usepackage{newfloat}
\usepackage{listings}
\DeclareCaptionStyle{ruled}{labelfont=normalfont,labelsep=colon,strut=off} % DO NOT CHANGE THIS
\floatstyle{ruled}
\newfloat{listing}{tb}{lst}{}
\floatname{listing}{Listing}

\usepackage{booktabs}

\title{Safety-Aligned Weights Are Not Enough: Refusal-Teacher-Guided Finetuning Enhances Safety and Utility under Harmful Finetuning Attacks}
\author{
     {Seokil Ham},
     {Yubin Choi},
     {Yujin Yang},
     {Younghun Kim}, 
     {Seungju Cho}, 
     {Changick Kim}\corresponding
}
\affiliations{
    \textsuperscript{\rm 1}Korea Advanced Institute of Science and Technology (KAIST), \\ 
    Daejeon, Republic of Korea \\
    \{gkatjrdlf, choibinbin, ujin.y, younghun1664, joyga, changick\}@kaist.ac.kr

}

\begin{document}

\maketitle

\begin{abstract}
While Finetuning-as-a-Service (FaaS) enables customization of Large Language Models (LLMs) using user data, this service is vulnerable to safety degradation when user data includes harmful prompts, a threat known as \textit{harmful finetuning attacks}.
To defend against this, prior work first constructs safety-aligned LLM and then finetunes the LLM on user data. However, we observe that the safety-aligned weights provide weak initialization for downstream task learning, leading to suboptimal safety and utility.
Motivated by this limitation, we shift the safe FaaS finetuning paradigm from \textit{finetuning safety-aligned weights} to \textit{finetuning base weights under explicit safety-teacher guidance}.
Specifically, we propose a \textbf{Refusal-Teacher (Ref-Teacher)-guided finetuning framework}. 
Our approach directly finetunes the base LLM under the guidance of a safety-aligned Ref-Teacher, which filters harmful prompts from user data and distills safety into the base LLM during finetuning.
Extensive experiments demonstrate that our paradigm effectively reduces harmful outputs while improving finetuning performance on user-specific tasks.
\end{abstract}


\section{Introduction}
% Recently, major AI service providers such as Google and OpenAI offer Finetuning-as-a-Service (FaaS), which enables users to upload custom datasets and adapt Large Language Models (LLMs) to their domain-specific task requirements. However, FaaS must prevent the malicious use of LLMs through safety alignment, even when users attempt to jailbreak the models via customization. These types of attacks, which inject harmful prompts into user data for finetuning, are called \textit{harmful finetuning attacks}. Several studies~\cite{qi2024fine,lermen2023lora,huang2025virus} have shown that finetuning on user data containing harmful content compromises the safety alignment, despite the LLMs being safety-aligned before finetuning. This vulnerability highlights the need to preserve safety while achieving high performance on user tasks in FaaS.

Recently, major AI service providers such as Google and OpenAI offer Finetuning-as-a-Service (FaaS), which enables users to adapt Large Language Models (LLMs) to their domain-specific tasks using custom datasets.
However, FaaS is vulnerable to \textit{harmful finetuning attacks}, where user data contains harmful prompts that degrade safety alignment during customization~\cite{lermen2023lora,qi2024fine,huang2025virusharmfulfinetuningattack}.
This vulnerability highlights a central challenge in FaaS: preserving safety while achieving strong performance on user-specific tasks.

% To mitigate these risks, prior work typically adopts a two-stage pipeline: (i) \textit{alignment stage}, where pre-trained LLMs are trained on safety-alignment data to prevent generating harmful responses, and (ii) \textit{finetuning stage}, where the safety-aligned models are finetuned on user data for user-specific downstream tasks.
% However, we observe that \textbf{this two-stage pipeline practice is suboptimal}. Safety-aligned models provide weak initialization for learning downstream tasks, resulting in limited task performance and degraded safety. A more effective alternative is to directly finetune the base model on both user data and safety-alignment data during finetuning stage, thereby enhancing downstream task performance while preserving safety. 
% Nevertheless, this strategy suffers from gradient conflicts between safety and utility objectives, which destabilize training and are further exacerbated when user data contains harmful prompts.  

To mitigate these risks, prior work typically adopts a two-stage pipeline: (i) an \textit{alignment stage}, where pre-trained LLMs are trained on safety-alignment data to prevent harmful responses, and (ii) a \textit{finetuning stage}, where the safety-aligned models are finetuned on user data for user-specific downstream tasks.
However, we observe that \textbf{this two-stage safety-aligned-weight finetuning paradigm is suboptimal}.
Safety-aligned weights can provide weak initialization for downstream adaptation, resulting in limited task performance and degraded safety after user finetuning.
In contrast, directly finetuning the base model on both user data and safety-alignment data offers a more effective alternative, as it preserves the base model's adaptability while enforcing safety during finetuning.
Nevertheless, naive joint finetuning introduces gradient conflicts between safety and utility objectives, which destabilize training and are further exacerbated when user data contains harmful prompts.

% \paragraph{Our Contributions.}
% Building on these observations, we propose a novel \textbf{Refusal-Teacher (Ref-Teacher)-guided finetuning framework} (Fig.~\ref{fig:main_figure}), which directly finetunes the base LLM on both user data and safety-alignment data under the guidance of a Ref-Teacher.  
% In our framework, the Ref-Teacher serves two complementary roles: 
% First, it performs \textbf{Alignment Distillation} by generating soft refusal labels that provide richer supervision and yield smoother loss surfaces, thereby mitigating gradient conflicts.  
% Second, it performs \textbf{Data Filtering} by identifying and removing harmful prompts from user data based on its refusal feature, ensuring robust conflict mitigation against harmful finetuning attacks.  
% Through these two complementary roles, our framework effectively alleviates gradient conflicts, resulting in improved safety and downstream task performance even under harmful finetuning attacks.
% Extensive experiments demonstrate that Ref-Teacher-guided finetuning consistently enhances user-specific task performance while preserving safety alignment, overcoming the limitations of prior two-stage pipelines and offering a practical solution for secure FaaS.

\paragraph{Our Contributions.}
Building on these observations, we introduce a paradigm shift for safe FaaS finetuning: from \textit{finetuning safety-aligned weights} to \textit{finetuning base weights under explicit safety-teacher guidance}.
To instantiate this paradigm, we propose a \textbf{Refusal-Teacher (Ref-Teacher)-guided finetuning framework} (Fig.~\ref{fig:main_figure}), which keeps the base LLM as an adaptable learner while dynamically enforcing safety during finetuning.
The Ref-Teacher serves two complementary roles.
First, it performs \textbf{Alignment Distillation} by generating soft refusal labels that provide richer supervision and yield smoother loss surfaces, thereby mitigating gradient conflicts.
Second, it performs \textbf{Data Filtering} by identifying and removing harmful prompts from user data based on its refusal feature, preventing harmful samples from destabilizing finetuning.
Through these mechanisms, our framework alleviates gradient conflicts and improves both safety and downstream task performance under harmful finetuning attacks.
Extensive experiments demonstrate that this paradigm effectively reduces harmful outputs while improving user-task performance across diverse harmful prompt ratios, user data sizes, downstream tasks, and model architectures.

\begin{table*}[t]
\centering
\scriptsize
\begin{adjustbox}{max width=\linewidth}
\begin{tabular}{l|cc|cc||cc|cc||cc|cc||cc|cc}
\toprule
\multirow{2}{*}{Methods} & \multicolumn{4}{c}{$p=0$} & \multicolumn{4}{c}{$p=0.1$} & \multicolumn{4}{c}{$p=0.3$} & \multicolumn{4}{c}{$p=0.5$} \\
\cmidrule{2-17}
& HS & FA & Freq (\%) & CS & HS & FA & Freq (\%) &  CS & HS & FA & Freq (\%) &  CS  & HS & FA & Freq (\%) &  CS  \\
\midrule
SA (1,000) $\rightarrow$ FT (1,000) & 4.9 & 42.8 & 3.37 & 0.574 & 48.1 & 43.4 & 3.54 & 0.551 & 78.2 & 40.2 & 3.54 & 0.531 & 79.8 & 42.7 & 3.45 & 0.525 \\ 
SA (5,000) $\rightarrow$ FT (1,000) & 3.3 & 41.3 & 4.27 & 0.540 & 22.8 & 41.9 & 3.86 & 0.525 & 61.7 & 39.4 & 4.71 & 0.500 & 71.1 & 39.7 & 4.30 & 0.487 \\ 
SA (10,000) $\rightarrow$ FT (1,000) & 2.2 & 41.1 & 3.29 & 0.549 & 16.2 & 39.9 & 3.93 & 0.524 & 57.3 & 39.1 & 4.03 & 0.501 & 71.3 & 37.1 & 4.13 & 0.525 \\ 
\cmidrule{1-17}
Base $\rightarrow$ SA (1,000) + FT (1,000) & 0.9 & 47.6 & 35.09 & 0.110 & 2.0 & 47.9 & 36.80 & 0.099 & 4.3 & 45.6 & 40.80 & 0.073 & 15.7 & 45.0 & 46.03 & 0.039 \\ 
\bottomrule
\end{tabular}
\end{adjustbox}
\vspace{-1mm}
\caption{Analysis under varying harmful prompt ratios $p$. We compare the conventional two-stage pipeline and direct base-model finetuning using Harmful Score (HS; lower is better), Fine-tuning Accuracy (FA; higher is better), gradient conflict frequency (Freq), and average gradient cosine similarity (CS). Freq denotes the percentage of gradients with negative cosine similarity. \textit{SA} and \textit{FT} denote safety alignment and finetuning, respectively, with data sizes in $(\cdot)$.}
\vspace{-2mm}
\label{tab:gradient conflict motivation}
\end{table*}

\section{Related Works}
\paragraph{Safety in Large Language Models.}
While LLMs can handle diverse queries, they remain vulnerable to harmful prompts that elicit unsafe outputs, such as weapon-making instructions~\cite{ji2023beavertails,chao2024jailbreakbench}. 
To mitigate this risk, LLMs are safety-aligned using datasets that pair harmful prompts with refusal responses~\cite{ouyang2022training,bianchi2024safety}. 
Nevertheless, safety-aligned LLMs are still susceptible to advanced jailbreaking~\cite{zou2023universal,liu2024autodan,chao2025jailbreaking}. 
Existing defenses broadly fall into training-free and training-based approaches. 
Training-free methods leverage representation-level differences between harmful and harmless inputs~\cite{xie2024gradsafe, hu2024gradient}.
Whereas, training-based methods enhance robustness through adversarial training by generating adversarial samples via latent space perturbations~\cite{xhonneux2024efficient,zou2024improving,yu2025robust,sheshadri2025latent}.
Recently, the \textit{refusal feature} has been identified as a vector encoding the rejection behavior of safety-aligned LLMs, and has been leveraged for both attack~\cite{arditi2024refusal} and defense~\cite{yu2025robust}.
Building on this insight, we utilize the refusal feature to detect harmful prompts within a novel framework for safe LLM finetuning.

\paragraph{Defending Harmful Finetuning Attacks.}
Harmful finetuning attacks inject harmful input-output pairs into training data, causing safety-aligned models to unlearn safe behaviors. Even small amounts of harmful data can significantly degrade safety~\cite{lermen2023lora,qi2024fine,zhan2024removing,hsu2024safe,he2024what,poppi2025towards,betley2025emergent,hsiung2026llm}, making robust defenses crucial for FaaS. Existing defenses target the alignment stage by learning weights robust to harmful updates~\cite{rosati2024representation,huang2024vaccine,huang2025booster,tamirisa2025tamper}, the finetuning stage by freezing safety-critical parameters or applying safety regularization~\cite{huang2024lisa,li2025safety,li2025salora,mukhoti2025fine,yang2026asft,ham2026jailbreak}, or the post-finetuning stage by editing safety-degrading weight shifts~\cite{hsu2024safe,huang2025antidote,yi2025nlsr,wang2026panacea}. In contrast, we identify a limitation of the standard pipeline and propose a Ref-Teacher-guided framework that directly finetunes the base model under a safety-aligned teacher to improve both safety and utility.

\section{Problem Setting}
\paragraph{Scenario \& Threat Model.}
In FaaS, AI providers aim to improve user-specific performance while preserving safety alignment. During the alignment stage, providers have access to $5{,}000$ harmful prompts paired with refusal responses and $5{,}000$ harmless prompts. During the finetuning stage, users submit customization data, where a fraction $p \in [0,1]$ consists of harmful prompts with harmful responses and the remaining $(1-p)$ consists of harmless prompts from the same dataset. Providers have no prior knowledge of whether user data contains harmful prompts or of its underlying distribution, making safe finetuning challenging.

\begin{figure*}[t]
    \centering
    \begin{subfigure}[b]{\textwidth}
        \centering
        \includegraphics[width=0.8\linewidth]{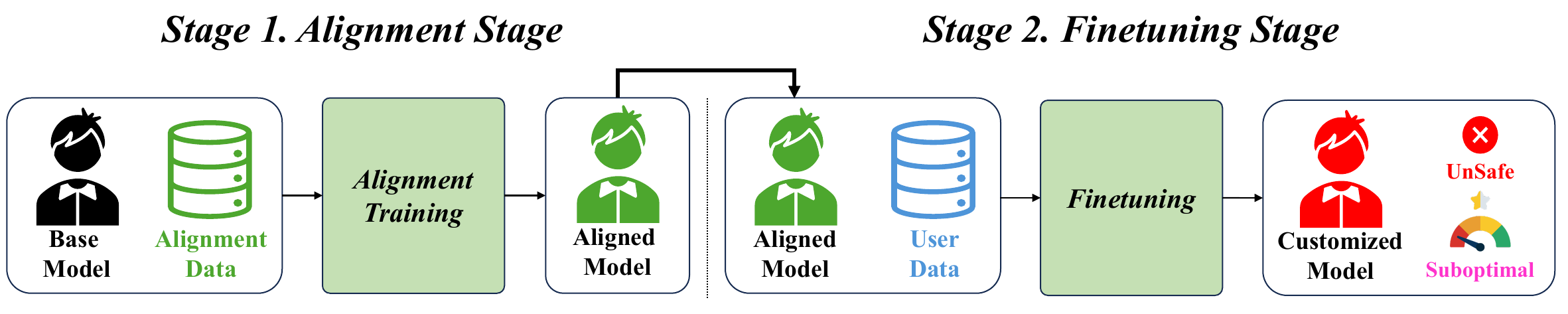}
        \caption{Baseline: Two-stage Pipeline}
        \label{fig:baseline}
    \end{subfigure}
    \begin{subfigure}[b]{\textwidth}
        \centering
        \includegraphics[width=0.9\linewidth]{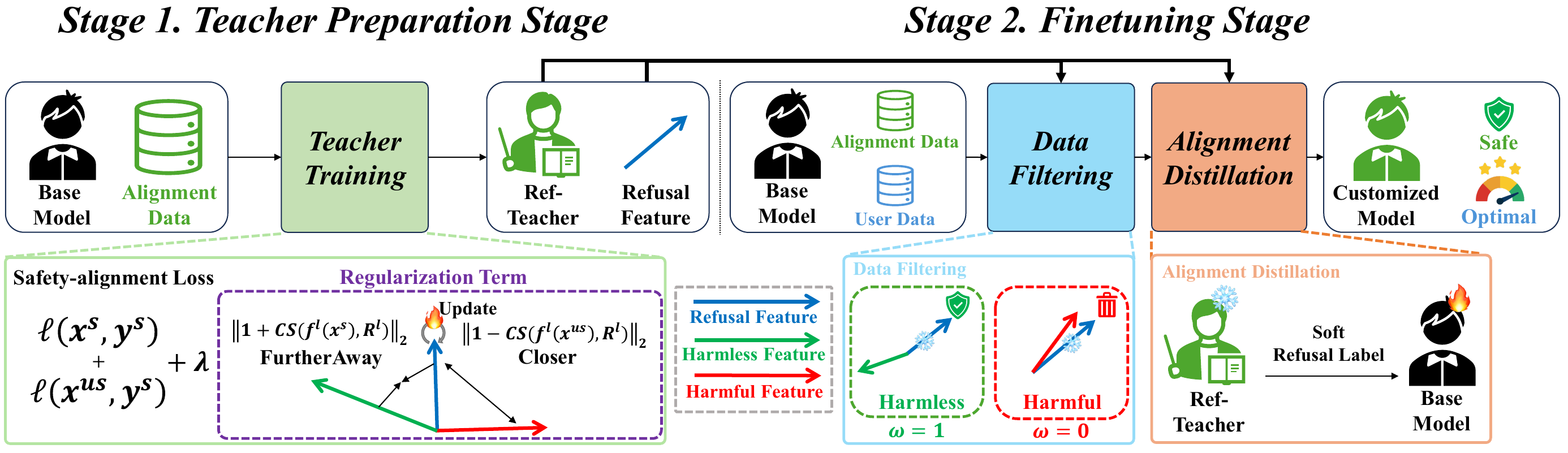}
        \caption{Ref-Teacher-guided Finetuning Framework}
        \vspace{-2mm}
        \label{fig:ours}
    \end{subfigure}
    \caption{Overview of our finetuning framework. (a) Baseline: The base LLM is first safety-aligned and then finetuned on user data. (b) Ours: The Ref-Teacher is trained with a dynamically updated refusal feature and then guides base-LLM finetuning on user and safety-alignment data via data filtering and alignment distillation.}
    \label{fig:main_figure}
    \vspace{-2mm}
\end{figure*}

% \section{Motivation}
% While prior work typically adopts a two-stage pipeline that first safety-aligns an LLM, and then finetunes the safety-aligned LLM on user data, we find this paradigm suboptimal. After an LLM is safety-aligned, its weights are biased toward safety objectives, weakening its effectiveness for learning downstream tasks compared to the base LLM. As a result, finetuning a safety-aligned model solely on user data yields limited downstream task performance and degraded safety alignment.
% In contrast, we observe that \textbf{directly finetuning the base LLM on both user data and safety-alignment data is more effective}. This strategy exploits the strong transferability of base LLMs for downstream learning while using safety-alignment data to counteract harmful finetuning.

\section{Motivation}
% Prior work typically follows a two-stage pipeline: first safety-aligning an LLM and then finetuning the safety-aligned model on user data. However, we find that this paradigm is suboptimal for both safety and utility. Once an LLM is safety-aligned, its weights are shifted toward safety objectives, which can weaken its ability to adapt to downstream tasks compared with the base LLM. Consequently, finetuning the safety-aligned model solely on user data often results in limited downstream task performance while also degrading its safety alignment.
% In contrast, we observe that \textbf{directly finetuning the base LLM on both user data and safety-alignment data is more effective}. This approach preserves the base LLM's task-learning capability while mitigating safety degradation from harmful finetuning.

% To validate this claim, we compare two finetuning strategies using Harmful Score (HS) and Finetuning Accuracy (FA) after finetuning (see Experiment Section for metric details): (i) finetuning safety-aligned LLMs solely on user data, and (ii) directly finetuning the base LLM on both user data and safety-alignment data.
% As shown in Table~\ref{tab:gradient conflict motivation}, safety-aligned models suffer from reduced downstream task performance, whereas finetuning the base LLM achieves both strong safety alignment and downstream task performance. 
% % Notably, even this simple direct finetuning strategy attains performance comparable to existing baselines.

We first examine whether the conventional two-stage pipeline is an optimal design choice for safe FaaS finetuning. In this pipeline, safety is encoded into model weights before user adaptation, and the resulting safety-aligned model is used as the initialization for downstream finetuning. However, this design can limit downstream adaptability. Since safety alignment shifts model parameters toward safety, the resulting weights may be less favorable for user-task adaptation.

Table~\ref{tab:gradient conflict motivation} compares this conventional pipeline with direct base-model finetuning on both user data and safety-alignment data. The results show that safety-aligned models often suffer from lower finetuning accuracy and can still lose safety after user finetuning. In contrast, directly finetuning the base LLM achieves a better safety and utility, suggesting that the base model provides a more effective initialization for user adaptation when safety data is also provided during finetuning.

\paragraph{Limitations.}
However, directly finetuning the base LLM on both user data and safety-alignment data introduces \textbf{gradient conflicts}, as the model must simultaneously optimize two distinct objectives.  
Following prior work~\cite{yu2020gradient, chen2020just}, we define gradient conflict as \textit{opposing update directions between gradients from different objectives}, indicated by negative cosine similarity.
Accordingly, we quantify this conflict by measuring cosine similarities between gradients from user and safety-alignment data across parameters over 300 training steps (Details in Supplementary Material).  
As shown in Table~\ref{tab:gradient conflict motivation}, directly finetuning the base LLM on user data and safety-alignment data leads to conflicts in more than 35\% of gradients, with harmful prompts in user data further exacerbating this issue.

This highlights a key challenge in moving beyond safety-aligned-weight finetuning: base-model finetuning preserves adaptability, but naive joint optimization creates safety-utility interference.
Therefore, safe FaaS finetuning requires a mechanism that can inject safety guidance during adaptation without undermining user-task learning.
% Motivated by this observation, we propose a {Ref-Teacher-guided finetuning framework}, which mitigates gradient conflicts through alignment distillation and data filtering, thereby stabilizing training and improving both safety and utility.

\section{Methodology}
% We propose a \textbf{Refusal-Teacher (Ref-Teacher)-guided finetuning framework} that directly finetunes the base LLM on both safety-alignment data and user data under the guidance of a Ref-Teacher via \textbf{alignment distillation} and \textbf{data filtering}. Unlike prior work that adopts the alignment stage, we introduce a \textbf{teacher preparation stage} to train the Ref-Teacher, followed by a finetuning stage where the base LLM is trained under its guidance. An overview of our framework and a comparison with prior work are shown in Fig.~\ref{fig:main_figure}.

To provide such a mechanism, we propose a \textbf{Refusal-Teacher (Ref-Teacher)-guided finetuning framework}, which keeps the base LLM adaptable while injecting safety guidance during finetuning.
Instead of using safety-aligned weights as the initialization, our framework directly finetunes the base LLM under a safety-aligned Ref-Teacher.
The Ref-Teacher guides this process via \textbf{alignment distillation}, which transfers safe behavior, and \textbf{data filtering}, which removes harmful user prompts.
The framework consists of a \textit{teacher preparation stage} for training the Ref-Teacher and a finetuning stage for adapting the base LLM under its guidance, as shown in Fig.~\ref{fig:main_figure}.

\subsection{Teacher Preparation Stage}
The goal of the teacher preparation stage is to train a safety-aligned teacher model that supports alignment distillation and data filtering during finetuning. To this end, we leverage the \textbf{refusal feature} during safety alignment to enable accurate discrimination between harmful and harmless prompts.

\begin{algorithm}[t]
\small
\caption{Training the Ref-Teacher Model}
\label{alg:reft_training}
\begin{algorithmic}[1]
\Require Harmful dataset $\mathcal{D}_{us}$, harmless dataset $\mathcal{D}_{s}$, update interval $C$, LoRA weights $W$, learning rate $\eta$
\Ensure Trained LoRA weights $W$, refusal feature $R^l$
\State $S_{us} \gets \emptyset$, $S_s \gets \emptyset$, $R^l \gets \varnothing$, $c \gets 0$
\While{not converged}
\State Sample mini-batches $\mathcal{B}_{us} \sim \mathcal{D}_{us}$ and $\mathcal{B}_s \sim \mathcal{D}_s$
\State $S_{us} \gets S_{us} \cup \mathcal{B}_{us}$, $S_s \gets S_s \cup \mathcal{B}_s$
\State $c \gets c + |\mathcal{B}_{us}|$
\If{$c \geq C$}
\State $\mu_{us}^l \gets \frac{1}{|S_{us}|} \sum_{x \in S_{us}} f^l(x)$
\State $\mu_s^l \gets \frac{1}{|S_s|} \sum_{x \in S_s} f^l(x)$
\State $R^l \gets \mu_{us}^l - \mu_s^l$
\State $S_{us} \gets \emptyset$, $S_s \gets \emptyset$, $c \gets 0$
\EndIf
\State Compute $\mathcal{L}_{\mathrm{align}}$
\If{$R^l \neq \varnothing$}
\State Compute $\mathcal{L}_{\mathrm{regul}}$ using $R^l$
\State $\mathcal{L}_{\mathrm{teacher}} \gets \mathcal{L}_{\mathrm{align}} + \lambda \mathcal{L}_{\mathrm{regul}}$
\Else
\State $\mathcal{L}_{\mathrm{teacher}} \gets \mathcal{L}_{\mathrm{align}}$
\EndIf
\State $W \gets W - \eta \nabla_W \mathcal{L}_{\mathrm{teacher}}$
\EndWhile
\State \Return $W$, $R^l$
\end{algorithmic}
\end{algorithm}

The refusal feature~\cite{arditi2024refusal} is a vector that encodes safety behavior, namely refusing harmful prompts while generating helpful responses for harmless ones.  
Formally, it is defined as the mean difference between feature of harmful and harmless prompts at a specific layer $l$ of the LLM.  
Let $x^s$ and $x^{us}$ denote safe and unsafe prompts, respectively, and let $f^l(\cdot)$ denote the feature of the last query token extracted from layer $l$. The refusal feature $R^l$ is computed as $R^l = \frac{1}{N_{us}} \sum^{N_{us}}_{i=1}{f^{l}(x^{us}_i)} - \frac{1}{N_{s}} \sum^{N_{s}}_{i=1}{f^{l}(x^{s}_i)}$
where $N_{us}$ and 
$N_{s}$ denote the number of unsafe and safe prompts, respectively.
As a result, harmful prompts exhibit high cosine similarity with $R^l$, while harmless prompts exhibit low similarity, enabling classification via a cosine similarity threshold.

Leveraging this property, we introduce the \textbf{Ref-Teacher}, a safety-aligned LLM that (i) generates soft refusal labels for alignment distillation and (ii) distinguishes harmful from harmless prompts effectively using its refusal feature for data filtering. To this end, we train the Ref-Teacher with a \textbf{safety-alignment loss} $\mathcal{L}_{\text{align}}$, defined as a supervised finetuning loss where harmful prompts are paired with refusal responses and harmless prompts are paired with helpful responses. 
This loss induces distinct behaviors across prompt types.

To further enhance the discrimination, we additionally introduce a \textbf{regularization loss} $\mathcal{L}_{\text{regul}}$ that enforces clearer separation between harmful and harmless prompt features based on the refusal feature. Specifically, this loss increases the cosine similarity between the refusal feature and harmful prompt features toward $1$, while decreasing its similarity with harmless prompt features toward $-1$. To avoid corrupting internal representations, its strength is controlled by $\lambda$.

The final teacher preparation objective combines safety-alignment loss and regularization loss:
\begin{equation}
\small
\label{eq: equation2}
\begin{split}
\mathcal{L}_{\text{align}} & = \frac{1}{N}\sum_{i=1}^{N} \big[ \ell(x^{s}_i, y^{s}_i) + \ell(x^{us}_i, y^{s}_i)\big],\\
\mathcal{L}_{\text{regul}} & = \frac{1}{N}\sum_{i=1}^{N} \big[ \lVert 1 + \mathrm{CS}(f^{\ell}(x^{s}_i), R^{\ell}) \rVert_2 \\ 
& \qquad \qquad + \lVert 1 - \mathrm{CS}(f^{\ell}(x^{us}_i), R^{\ell}) \rVert_2 \big], \\
\mathcal{L}_{\text{teacher}} & = \mathcal{L}_{\text{align}} + \lambda \cdot \mathcal{L}_{\text{regul}}, 
\end{split}
\end{equation}
where $\ell(\cdot, \cdot)$ is the cross-entropy loss, $CS(\cdot, \cdot)$ is cosine similarity, $y^s$ is the safe (helpful or refusal) responses, respectively, and $N$ is the number of training samples.  
Eq.~\ref{eq: equation2} enables the Ref-Teacher to both generate appropriate refusal responses for harmful prompts and reliably distinguish harmful from harmless inputs using refusal features.

In addition, we assume a setting where a pre-aligned model is unavailable, making it impossible to extract the refusal feature in advance.
To address this, we dynamically update the refusal feature during training at fixed intervals by accumulating harmful and harmless prompts into $S_{us}$ and $S_s$. Before the first update, we disable the regularization loss because the refusal feature is unreliable. This dynamic update strategy eliminates the need for a separate alignment stage, enabling the model to compute the refusal feature and learn discriminative representations within a single training process. The Ref-Teacher training procedure is summarized in Alg.~\ref{alg:reft_training}.

\begin{table*}[t]
\centering
\scriptsize
\renewcommand{\arraystretch}{0.9}
\begin{adjustbox}{max width=\textwidth}
\begin{tabular}{l|ccccc|ccccc}
\toprule
\multirow{2}{*}{Methods} & \multicolumn{5}{c}{Harmful Score ($\downarrow$)} & \multicolumn{5}{c}{Finetuning Accuracy ($\uparrow$)} \\
\cmidrule(lr){2-6} \cmidrule(lr){7-11}
& $p=0$ & $p=0.1$ & $p=0.3$ & $p=0.5$ & Average & $p=0$ & $p=0.1$ & $p=0.3$ & $p=0.5$ & Average \\
\midrule
SFT&$2.2_{\pm0.1}$&$16.2_{\pm0.4}$&$57.3_{\pm0.6}$&$71.3_{\pm0.6}$&$36.8_{\pm0.2}$&$41.1_{\pm0.0}$&$39.9_{\pm0.6}$&$39.1_{\pm0.2}$&$37.1_{\pm0.6}$&$39.3_{\pm0.2}$\\
Repnoise&$2.7_{\pm0.4}$&$29.9_{\pm0.6}$&$67.0_{\pm5.1}$&$75.7_{\pm3.1}$&$43.8_{\pm2.0}$&$37.4_{\pm0.3}$&$37.0_{\pm1.2}$&$36.3_{\pm0.7}$&$36.0_{\pm1.4}$&$36.7_{\pm0.2}$\\
Vaccine&$1.3_{\pm0.2}$&$5.4_{\pm0.7}$&$35.0_{\pm0.3}$&$57.5_{\pm0.4}$&$24.8_{\pm0.2}$&$22.9_{\pm0.5}$&$23.2_{\pm1.0}$&$21.7_{\pm0.3}$&$20.3_{\pm0.4}$&$22.0_{\pm0.3}$\\
Booster&$2.3_{\pm0.1}$&$5.9_{\pm0.2}$&$65.1_{\pm0.3}$&$75.0_{\pm0.6}$&$37.1_{\pm0.2}$&$44.5_{\pm0.5}$&$44.0_{\pm0.9}$&$44.4_{\pm0.6}$&$43.5_{\pm0.6}$&$44.1_{\pm0.4}$\\
LDIFS&$1.0_{\pm0.2}$&$4.1_{\pm0.7}$&$7.1_{\pm0.2}$&$14.7_{\pm0.3}$&$6.7_{\pm0.2}$&$18.0_{\pm0.5}$&$16.7_{\pm0.9}$&$15.5_{\pm0.6}$&$15.4_{\pm0.6}$&$16.4_{\pm0.4}$\\
SafeInstruct&$1.0_{\pm0.1}$&$6.4_{\pm0.6}$&$30.9_{\pm2.6}$&$53.7_{\pm4.3}$&$23.0_{\pm1.8}$&$44.4_{\pm0.2}$&$42.2_{\pm0.8}$&$41.3_{\pm0.5}$&$39.8_{\pm0.2}$&$41.9_{\pm0.4}$\\
Lisa&$1.4_{\pm0.2}$&$5.3_{\pm0.1}$&$25.9_{\pm1.5}$&$49.2_{\pm0.7}$&$20.4_{\pm0.5}$&$38.3_{\pm0.7}$&$38.9_{\pm0.9}$&$37.8_{\pm0.9}$&$36.2_{\pm0.5}$&$37.8_{\pm0.3}$\\
Antidote&$4.5_{\pm0.2}$&$21.4_{\pm0.2}$&$54.0_{\pm0.3}$&$65.4_{\pm0.7}$&$36.4_{\pm0.1}$&$47.0_{\pm0.8}$&$46.2_{\pm0.3}$&$44.5_{\pm0.8}$&$42.6_{\pm0.6}$&$45.1_{\pm0.2}$\\
\midrule
SFT*&$1.8_{\pm0.0}$&$1.9_{\pm0.2}$&$1.5_{\pm0.1}$&$1.5_{\pm0.2}$&$1.7_{\pm0.1}$&$43.7_{\pm0.5}$&$42.5_{\pm0.2}$&$41.1_{\pm0.7}$&$38.1_{\pm1.0}$&$41.3_{\pm0.4}$\\
Repnoise*&$2.5_{\pm0.1}$&$2.7_{\pm0.0}$&$2.1_{\pm0.2}$&$1.3_{\pm0.2}$&$2.2_{\pm0.1}$&$37.3_{\pm0.8}$&$37.6_{\pm0.9}$&$36.2_{\pm0.5}$&$32.9_{\pm0.5}$&$36.0_{\pm0.1}$\\
Vaccine*&$1.5_{\pm0.1}$&$1.2_{\pm0.1}$&$1.1_{\pm0.1}$&$1.2_{\pm0.1}$&$1.3_{\pm0.1}$&$22.9_{\pm0.3}$&$22.1_{\pm0.4}$&$16.8_{\pm0.2}$&$11.0_{\pm0.4}$&$18.2_{\pm0.0}$\\
Booster*&$1.3_{\pm0.2}$&$1.3_{\pm0.0}$&$1.2_{\pm0.0}$&$1.5_{\pm0.0}$&$1.3_{\pm0.0}$&$43.9_{\pm0.7}$&$43.6_{\pm0.0}$&$42.0_{\pm0.6}$&$42.8_{\pm0.1}$&$43.1_{\pm0.3}$\\
LDIFS*&$1.0_{\pm0.1}$&$1.1_{\pm0.2}$&$1.0_{\pm0.1}$&$\textbf{0.9}_{\pm0.2}$&$1.0_{\pm0.0}$&$17.5_{\pm1.1}$&$16.7_{\pm0.1}$&$18.0_{\pm0.1}$&$17.2_{\pm0.9}$&$17.3_{\pm0.2}$\\
SafeInstruct*&$1.0_{\pm0.1}$&$1.1_{\pm0.1}$&$1.1_{\pm0.1}$&$1.1_{\pm0.1}$&$1.1_{\pm0.1}$&$44.1_{\pm0.5}$&$43.1_{\pm0.5}$&$40.7_{\pm0.9}$&$38.1_{\pm0.3}$&$41.5_{\pm0.5}$\\
Lisa*&$1.3_{\pm0.3}$&$1.3_{\pm0.1}$&$1.2_{\pm0.1}$&$1.3_{\pm0.1}$&$1.3_{\pm0.0}$&$38.6_{\pm0.5}$&$38.0_{\pm1.7}$&$35.7_{\pm0.3}$&$33.9_{\pm1.5}$&$36.5_{\pm0.3}$\\
Antidote*&$1.2_{\pm0.1}$&$1.3_{\pm0.1}$&$1.1_{\pm0.2}$&$1.1_{\pm0.1}$&$1.2_{\pm0.0}$&$46.7_{\pm0.3}$&$46.1_{\pm0.6}$&$43.8_{\pm0.2}$&$41.4_{\pm0.9}$&$44.5_{\pm0.1}$\\
\cmidrule{1-11}
% Deep Ignorance &$1.9_{\pm0.2}$&$14.4_{\pm0.6}$&$47.1_{\pm0.6}$&$65.2_{\pm0.5}$&$32.2_{\pm0.3}$&$43.1_{\pm0.5}$&$42.4_{\pm0.2}$&$42.1_{\pm0.4}$&$42.2_{\pm0.6}$&$42.5_{\pm0.3}$\\
% \cmidrule{1-11}
Ours (Base) &$\textbf{0.7}_{\pm0.1}$&$1.2_{\pm0.7}$&$3.3_{\pm0.3}$&$7.3_{\pm0.4}$&$3.1_{\pm0.2}$&${46.9}_{\pm0.8}$&$47.9_{\pm0.8}$&$\textbf{46.9}_{\pm0.3}$&$44.4_{\pm0.4}$&$46.5_{\pm0.2}$\\
Ours (Ref-Teacher) &$0.9_{\pm0.3}$&$\textbf{1.0}_{\pm0.5}$&$\textbf{0.6}_{\pm0.1}$&$\textbf{0.9}_{\pm0.3}$&$\textbf{0.9}_{\pm0.2}$&$\textbf{48.8}_{\pm0.5}$&$\textbf{49.0}_{\pm0.5}$&$45.5_{\pm0.9}$&$\textbf{44.8}_{\pm0.5}$&$\textbf{47.0}_{\pm0.1}$\\
\bottomrule
\end{tabular}
\end{adjustbox}
\vspace{-1mm}
\caption{Performance under varying harmful prompt ratios $p$. We report HS and FA to assess the safety-utility trade-off as the proportion of harmful user data increases. Methods marked with * use LLaMAGuard3-8B-based data filtering.}
\label{tab:harmful_ratio_filtered}
\end{table*}

\begin{table*}[!t]
\centering
\scriptsize
\renewcommand{\arraystretch}{0.9}
\begin{adjustbox}{max width=\textwidth}
\begin{tabular}{l|ccccc|ccccc}
\toprule
\multirow{2}{*}{Methods} & \multicolumn{5}{c}{Harmful Score ($\downarrow$)} & \multicolumn{5}{c}{Finetuning Accuracy ($\uparrow$)} \\
\cmidrule(lr){2-6} \cmidrule(lr){7-11}
& n=1000 & n=1500 & n=2000 & n=2500 & Average & n=1000 & n=1500 & n=2000 & n=2500 & Average \\
\midrule
SFT & $16.2_{\pm0.4}$ & $38.7_{\pm0.6}$ & $55.7_{\pm0.7}$ & $63.2_{\pm0.6}$ & $44.3_{\pm0.4}$ & $39.9_{\pm0.6}$ & $43.5_{\pm0.6}$ & $46.8_{\pm2.0}$ & $46.4_{\pm1.2}$ & $44.5_{\pm1.0}$\\
Repnoise & $29.9_{\pm0.6}$ & $45.9_{\pm4.1}$ & $60.4_{\pm1.2}$ & $68.1_{\pm4.2}$ & $51.1_{\pm2.4}$ & $37_{\pm1.2}$ & $40.8_{\pm0.4}$ & $42.1_{\pm1.4}$ & $43.6_{\pm0.1}$ & $40.9_{\pm0.6}$\\
Vaccine & $5.4_{\pm0.7}$ & $16.3_{\pm3.0}$ & $30.3_{\pm3.5}$ & $40.4_{\pm4.1}$ & $23.1_{\pm2.5}$ & $23.2_{\pm1.0}$ & $28.0_{\pm0.4}$ & $32.7_{\pm1.2}$ & $34.1_{\pm0.5}$ & $29.5_{\pm0.1}$\\
Booster & $5.9_{\pm0.2}$ & $14.5_{\pm4.3}$ & $43.5_{\pm4.1}$ & $59.3_{\pm2.9}$ & $30.8_{\pm2.8}$ & $44.0_{\pm0.9}$ & $45.3_{\pm0.0}$ & $47.7_{\pm0.6}$ & $49.0_{\pm0.5}$ & $46.5_{\pm0.2}$\\
LDIFS & $4.1_{\pm0.7}$ & $3.9_{\pm1.6}$ & $3.5_{\pm1.1}$ & $4.5_{\pm1.4}$ & $4.0_{\pm1.0}$ & $16.7_{\pm0.8}$ & $16.9_{\pm0.2}$ & $17.7_{\pm0.8}$ & $17.6_{\pm0.7}$ & $17.2_{\pm0.5}$\\
SafeInstruct & $6.6_{\pm0.3}$ & $9.8_{\pm1.6}$ & $17.4_{\pm9.9}$ & $11.3_{\pm2.3}$ & $11.3_{\pm2.2}$ & $42.5_{\pm0.9}$ & $45.8_{\pm0.6}$ & $47.0_{\pm0.9}$ & $46.3_{\pm0.7}$ & $45.4_{\pm0.1}$\\
Lisa & $5.3_{\pm0.1}$ & $7.2_{\pm0.9}$ & $8.8_{\pm1.4}$ & $11.3_{\pm1.6}$ & $8.1_{\pm0.9}$ & $38.9_{\pm0.9}$ & $37.9_{\pm0.2}$ & $40.9_{\pm0.6}$ & $42.2_{\pm0.5}$ & $40.0_{\pm0.3}$\\
Antidote & $12.9_{\pm7.6}$ & $22.3_{\pm12.8}$ & $32.8_{\pm15.3}$ & $37.1_{\pm15.4}$ & $26.3_{\pm12.8}$ & $47.0_{\pm1.1}$ & $48.7_{\pm0.8}$ & $48.8_{\pm0.7}$ & $48.9_{\pm0.9}$ & $48.4_{\pm0.3}$\\
\midrule
SFT* & $3.4_{\pm2.8}$ & $1.8_{\pm0.2}$ & $2.4_{\pm0.3}$ & $4.4_{\pm0.2}$ & $3.0_{\pm0.7}$ & $42.5_{\pm0.2}$ & $43.6_{\pm0.7}$ & $45.9_{\pm0.2}$ & $44.4_{\pm0.7}$ & $43.9_{\pm0.4}$\\
Repnoise* & $6.2_{\pm6.1}$ & $2.8_{\pm0.3}$ & $3.9_{\pm0.1}$ & $5.2_{\pm0.1}$ & $4.6_{\pm1.5}$ & $37.2_{\pm0.5}$ & $40.4_{\pm1.0}$ & $42.3_{\pm0.4}$ & $42.5_{\pm0.2}$ & $40.6_{\pm0.1}$\\
Vaccine* & $1.4_{\pm0.4}$ & $1.3_{\pm0.1}$ & $1.3_{\pm0.1}$ & $1.6_{\pm0.2}$ & $1.4_{\pm0.2}$ & $22.4_{\pm0.5}$ & $26.6_{\pm0.6}$ & $29.6_{\pm0.4}$ & $33.3_{\pm0.9}$ & $28.0_{\pm0.5}$\\
Booster* & $1.9_{\pm1.1}$ & $1.4_{\pm0.1}$ & $2.3_{\pm0.1}$ & $2.0_{\pm0.1}$ & $1.9_{\pm0.3}$ & $43.6_{\pm0.1}$ & $45.2_{\pm0.2}$ & $45.2_{\pm0.1}$ & $48.3_{\pm0.5}$ & $45.6_{\pm0.2}$\\
LDIFS* & $1.6_{\pm0.9}$ & $1.1_{\pm0.3}$ & $1.6_{\pm0.9}$ & $1.8_{\pm0.9}$ & $1.5_{\pm0.7}$ & $16.9_{\pm0.4}$ & $17.9_{\pm0.7}$ & $18.5_{\pm0.6}$ & $18.2_{\pm1.4}$ & $17.9_{\pm0.5}$\\
SafeInstruct* & $1.1_{\pm0.1}$ & $1.2_{\pm0.1}$ & $1.1_{\pm0.2}$ & $1.0_{\pm0.1}$ & $1.1_{\pm0.0}$ & $43.1_{\pm0.5}$ & $44.8_{\pm0.3}$ & $45.8_{\pm0.9}$ & $46.5_{\pm0.6}$ & $45.0_{\pm0.3}$\\
Lisa* & $1.5_{\pm0.4}$ & $1.1_{\pm0.1}$ & $\textbf{1.0}_{\pm0.1}$ & $1.2_{\pm0.2}$ & $1.2_{\pm0.1}$ & $37.5_{\pm1.5}$ & $40.1_{\pm0.6}$ & $42.6_{\pm0.4}$ & $42.7_{\pm0.1}$ & $40.7_{\pm0.5}$\\
Antidote* & $1.3_{\pm0.1}$ & $1.2_{\pm0.1}$ & $1.5_{\pm0.2}$ & $1.5_{\pm0.2}$ & $1.4_{\pm0.1}$ & $46.1_{\pm0.6}$ & $47.1_{\pm0.9}$ & $47.4_{\pm0.4}$ & $48.4_{\pm0.3}$ & $47.2_{\pm0.2}$\\
% \cmidrule{1-11}
% Deep Ignorance & $14.4_{\pm0.6}$ & $32.4_{\pm1.7}$ & $46.0_{\pm0.9}$ & $50.0_{\pm6.1}$ & $35.7_{\pm1.3}$ & $42.4_{\pm0.2}$ & $44.4_{\pm0.4}$ & $44.6_{\pm0.3}$ & $44.8_{\pm0.7}$ & $44.1_{\pm0.1}$\\
% Pharmacist*~(\cite{liu2025pharmacist})  & 60.7 & 59.8 & 59.2 & 58.5 & 59.6 & 37.6 & 37.0 & 35.6 & 31.9 & 35.5 \\
\cmidrule{1-11}
% Base $\rightarrow$ SA (1,000) + FT (1,000)  & {0.9} & 2.0 & 4.3 & 15.7 & 5.7 & {47.6} & {47.9} & \textbf{45.6} & {45.0} & {46.5} \\
Ours (Base) & $1.2_{\pm0.7}$ & $1.7_{\pm0.2}$ & $1.7_{\pm0.4}$ & $1.0_{\pm0.1}$ & $1.4_{\pm0.1}$ & $47.9_{\pm0.8}$ & $46.8_{\pm0.4}$ & $48.2_{\pm0.3}$ & $50.2_{\pm1.5}$ & $48.3_{\pm0.5}$\\
Ours (Ref-Teacher) & $\textbf{1.0}_{\pm0.5}$ & $\textbf{0.8}_{\pm0.1}$ & $1.1_{\pm0.7}$ & $\textbf{0.9}_{\pm0.1}$ & $\textbf{1.0}_{\pm0.2}$ & $\textbf{49.0}_{\pm0.5}$ & $\textbf{50.2}_{\pm0.1}$ & $\textbf{51.2}_{\pm0.8}$ & $\textbf{51.8}_{\pm0.3}$ & $\textbf{50.5}_{\pm0.3}$\\
\bottomrule
\end{tabular}
\end{adjustbox}
\vspace{-1mm}
\caption{Performance under varying amounts of user data. $n$ denotes the number of user samples. The results evaluate scalability in preserving safety while improving user task performance. Methods marked with * use LLaMAGuard3-8B-based data filtering.}
\label{tab:num_samples_filtered}
\end{table*}

\begin{table*}[!t]
\centering
\renewcommand{\arraystretch}{0.9}
\scriptsize
\begin{adjustbox}{max width=\textwidth}
\begin{tabular}{l|cc|cc|cc|cc|cc}
\toprule
\multirow{2}{*}{Methods} & \multicolumn{2}{c}{GSM8K} & \multicolumn{2}{c}{SST2} & \multicolumn{2}{c}{AGNEWS} & \multicolumn{2}{c}{AlpacaEval} & \multicolumn{2}{c}{Average} \\
\cmidrule{2-11}
 & HS $\downarrow$ & FA $\uparrow$ & HS $\downarrow$ & FA $\uparrow$ & HS $\downarrow$ & FA $\uparrow$ & HS $\downarrow$ & FA $\uparrow$ & HS $\downarrow$ & FA $\uparrow$ \\
\midrule
SFT & $19.5_{\pm2.5}$ & $41.2_{\pm0.6}$ & {$41.7_{\pm7.1}$} & $93.3_{\pm0.1}$ & {$26.8_{\pm1.6}$} & $82.6_{\pm0.3}$ & 23.7 & 32.7 & 20.4 & 49.9 \\
Repnoise & $29.9_{\pm0.6}$ & $37.0_{\pm1.2}$ & $58.0_{\pm4.3}$ & $93.1_{\pm0.2}$ & $54.3_{\pm3.7}$ & $85.5_{\pm0.8}$ & 45.4 & 29.3 & 39.48 & 49.1 \\
Vaccine & $5.4_{\pm0.7}$ & $23.2_{\pm1.0}$ & $31.0_{\pm4.1}$ & $89.4_{\pm0.6}$ & $26.2_{\pm2.9}$ & $82.9_{\pm0.3}$ & 55.8 & 14.4 & 25.2 & 42.4 \\
Booster & $5.9_{\pm0.2}$ & $44.0_{\pm0.9}$ & $9.3_{\pm0.1}$ & $93.4_{\pm0.1}$ & $5.0_{\pm0.3}$ & $84.5_{\pm0.7}$ & 29.4 & 34.0 & 10.0 & 51.3 \\
LDIFS & $4.1_{\pm0.7}$ & $16.7_{\pm0.8}$ & $10.6_{\pm3.5}$ & $87.8_{\pm2.4}$ & {$9.2_{\pm2.9}$} & $69.3_{\pm1.6}$ & 5.7 & 33.7 & 7.4 & 42.5 \\
SafeInstruct & $6.6_{\pm0.3}$ & $42.5_{\pm0.9}$ & {$7.9_{\pm0.8}$} & $93.2_{\pm0.2}$ & {$6.8_{\pm0.5}$} & $86.4_{\pm0.1}$ & 33.0 & 59.0 & 11.3 & 56.0 \\
Lisa & $5.3_{\pm0.1}$ & $38.9_{\pm0.9}$ & {$18.3_{\pm2.7}$} & $92.8_{\pm0.5}$ & {$13.3_{\pm1.5}$} & $85.6_{\pm0.9}$ & 10.1 & 29.6 & 10.3 & 49.2 \\
Antidote & $12.9_{\pm7.6}$ & $47.0_{\pm1.1}$ & {$23.2_{\pm0.3}$} & $93.1_{\pm0.0}$ & $16.8_{\pm0.2}$ & $85.7_{\pm0.3}$ & 8.7 & 14.7 & 10.1 & 20.2 \\
\midrule
SFT* & $3.4_{\pm2.8}$ & $41.7_{\pm1.2}$ & {$\textbf{0.7}_{\pm0.1}$} & $90.8_{\pm0.4}$ & {$\textbf{0.6}_{\pm0.1}$} & $78.8_{\pm0.3}$ & 46.5 & 67.9 & 12.8 & 69.8 \\
Repnoise* & $6.2_{\pm6.1}$ & $37.2_{\pm0.5}$ & $0.9_{\pm0.1}$ & $92.6_{\pm0.1}$ & $0.9_{\pm0.1}$ & $79.4_{\pm0.5}$ & 41.1 & 66.2 & 12.3 & 68.9 \\
Vaccine* & $1.4_{\pm0.4}$ & $22.4_{\pm0.5}$ & $1.2_{\pm0.0}$ & $86.9_{\pm0.2}$ & $1.0_{\pm0.1}$ & $77.9_{\pm0.1}$ & 57.3 & 39.0 & 15.2 & 56.6 \\
Booster* & $1.9_{\pm1.1}$ & $43.6_{\pm0.1}$ & $2.0_{\pm0.0}$ & $93.7_{\pm0.0}$ & $1.5_{\pm0.0}$ & $82.4_{\pm0.0}$ & 57.9 & 65.7 & 15.8 & 71.4 \\
LDIFS* & $1.6_{\pm0.9}$ & $16.9_{\pm0.4}$ & $0.9_{\pm0.1}$ & $85.9_{\pm0.2}$ & {$\textbf{0.6}_{\pm0.1}$} & $66.0_{\pm0.7}$ & 3.1 & 44.0 & 1.5 & 53.2 \\
SafeInstruct* & $1.1_{\pm0.1}$ & $43.1_{\pm0.5}$ & {$\textbf{0.7}_{\pm0.1}$} & $91.7_{\pm0.2}$ & {$\textbf{0.6}_{\pm0.1}$} & $76.1_{\pm0.8}$ & 4.9 & 64.1 & 1.8 & 68.7 \\
Lisa* & $1.5_{\pm0.4}$ & $37.5_{\pm1.5}$ & {$\textbf{0.7}_{\pm0.1}$} & $91.0_{\pm0.1}$ & {$\textbf{0.6}_{\pm0.1}$} & $79.3_{\pm0.0}$ & 8.6 & 61.2 & 2.8 & 67.3 \\
Antidote* & $1.3_{\pm0.1}$ & $46.1_{\pm0.6}$ & {$\textbf{0.7}_{\pm0.1}$} & $81.5_{\pm0.2}$ & $0.7_{\pm0.1}$ & $63.3_{\pm0.8}$ & 24.0 & 69.2 & 6.7 & 65.0 \\
% \cmidrule{1-11}
% Deep Ignorance & $14.4_{\pm0.6}$ & $42.4_{\pm0.2}$ & $29.2_{\pm0.5}$ & $94.1_{\pm0.1}$ & $19.0_{\pm0.4}$ & {$\textbf{86.8}_{\pm0.2}$} & 64.2 & 62.7 & 31.7 & 71.5 \\
% Pharmacist*~(\cite{liu2025pharmacist}) & $59.9_{\pm0.8}$ & $37.5_{\pm0.5}$ & $58.4_{\pm0.8}$ & $93.3_{\pm0.1}$ & $56.8_{\pm0.5}$ & $85.8_{\pm0.2}$ & 62.9 & 63.8 & 59.5 & 70.1 \\
\cmidrule{1-11}
Ours (Base) & $1.2_{\pm0.7}$ & $47.9_{\pm0.8}$ & $2.6_{\pm2.9}$ & $94.7_{\pm0.1}$ & $2.3_{\pm2.8}$ & $84.6_{\pm0.8}$ & 7.3 & 64.4 & 3.3 & 72.9 \\
Ours (Ref-Teacher) & {$\textbf{1.0}_{\pm0.5}$} & {$\textbf{49.0}_{\pm0.5}$} & $1.1_{\pm0.2}$ & {$\textbf{95.0}_{\pm0.4}$} & $0.9_{\pm0.3}$ & $85.8_{\pm0.3}$ & \textbf{2.4} & \textbf{72.1} & \textbf{1.3} & \textbf{75.5} \\
\bottomrule
\end{tabular}
\end{adjustbox}
\vspace{-1mm}
\caption{Performance across downstream user-task datasets. Results on GSM8K, SST2, AGNEWS, and AlpacaEval evaluate robustness to different harmless data distributions. Methods marked with * use LLaMAGuard3-8B-based data filtering.}
\label{tab:dataset_filtered}
\vspace{-1mm}
\end{table*}

\subsection{Finetuning Stage}
In the finetuning stage, the Ref-Teacher is frozen and serves as a teacher for two complementary roles: (i) alignment distillation and (ii) data filtering. This approach mitigates gradient conflicts that arise during finetuning, allowing the base LLM to {simultaneously} learn user-specific tasks and safety alignment with strong performance on both.

\paragraph{Alignment Distillation.}  
Knowledge distillation is a common approach for mitigating gradient conflicts in multi-objective learning, as soft labels from a teacher provide richer supervision and yield smoother loss surfaces than hard labels~\cite{hinton2015distillingknowledgeneuralnetwork,furlanello2018born,muller2019does,yuan2020revisiting}.  
Building on this, we introduce \textbf{alignment distillation} to guide the base LLM in directly learning both user-specific tasks and safety alignment.  
Specifically, the Ref-Teacher generates soft refusal labels, and the base LLM is trained with (i) a supervised loss on user data and (ii) a KL-divergence loss on safety-alignment data to match the Ref-Teacher's safe behavior.  
This stabilizes training by reducing gradient conflicts, resulting in safe and appropriate responses for both harmful and user-specific inputs. To ensure the reliability of the soft refusal labels, we reuse the safety-alignment data from the teacher preparation stage, on which the Ref-Teacher has already been trained. 
This strategy eliminates the need for an additional safety-alignment dataset during finetuning.

\paragraph{Data Filtering.} 
While alignment distillation mitigates gradient conflicts between safety and user-specific task objectives, it alone cannot prevent these conflicts from being exacerbated by harmful finetuning attacks. Therefore, we adopt data filtering as a complementary solution. In our framework, the Ref-Teacher distinguishes harmful from harmless prompts in user data using its refusal feature. Specifically, we compute the cosine similarity between the refusal feature $R^l$ and each prompt feature $f^l(x_i)$; prompts with similarity exceeding a threshold $\tau$ are classified as harmful, and the remaining prompts as harmless. This filtering mechanism is formulated as a binary indicator $\omega_i$:
\begin{equation}
\small
\label{eq: filtering}
    \omega_i = 
    \begin{cases}
      0, & \text{if} \quad CS(R^l, f^l(x_i)) > \tau \\
      1, & otherwise \\
    \end{cases}.
\end{equation}
In Eq.~\ref{eq: filtering}, prompts classified as harmful are excluded from the supervised finetuning loss by setting $\omega_i = 0$, thereby reducing gradient conflicts that destabilize training.

\paragraph{Overall Objective.}  Our Ref-Teacher-guided finetuning framework integrates two complementary roles of the Ref-Teacher by combining a filtered supervised loss that learns user-specific tasks with an alignment distillation loss that transfers safety behavior from the Ref-Teacher to the base LLM. The overall finetuning objective is defined as:
\begin{equation}
\small
\begin{split}
    \mathcal{L}_\text{ft} &= \frac{1}{N_\text{user}} \sum^{N_\text{user}}_{i=1} \omega_i \cdot \ell(x_i, y_i) \\
    &\quad + \alpha T^2 \cdot \frac{1}{N_\text{safety}} \sum^{N_\text{safety}}_{i=1} \mathrm{KL}(p_{t,i}^T \,||\, p_{s,i}^T),
\end{split}
\end{equation}
where $N_\text{user}$ and $N_\text{safety}$ denote the numbers of user and safety-alignment samples, respectively. The first term $\ell(x_i,y_i)$ is a cross-entropy loss on user data, weighted by the filtering indicator $\omega_i$ to learn user-specific tasks. The second term is the alignment distillation loss on safety-alignment data, where $\mathrm{KL}$ measures the KL-divergence between the temperature-scaled output distributions of the Ref-Teacher, $p_{t,i}^T$, and the base LLM, $p_{s,i}^T$, with temperature $T$. The hyperparameter $\alpha$ controls the relative weight of the distillation term.

\begin{table*}[!t]
    \centering
    \scriptsize
    \renewcommand{\arraystretch}{0.9}
        % \begin{adjustbox}{max width=\textwidth}
        \begin{tabular}{l|cc|cc|cc|cc}
        \toprule
        \multirow{2.5}{*}{Methods} & \multicolumn{2}{c}{Llama3-8B} & \multicolumn{2}{c}{Gemma2-9B} & \multicolumn{2}{c}{Qwen2-7B} & \multicolumn{2}{c}{Average} \\
        \cmidrule{2-9}
         & HS $\downarrow$ & FA $\uparrow$ & HS $\downarrow$ & FA $\uparrow$ & HS $\downarrow$ & FA $\uparrow$ & HS $\downarrow$ & FA $\uparrow$ \\
        \midrule
        SFT & $19.5_{\pm2.5}$ & $41.2_{\pm0.6}$ & $26.1_{\pm0.3}$ & $56.8_{\pm0.4}$ & $11.3_{\pm0.8}$ & $60.0_{\pm1.1}$ & $19.0_{\pm0.9}$ & $52.7_{\pm0.6}$  \\
        Repnoise & $29.9_{\pm0.6}$ & $37.0_{\pm1.2}$ & $23.8_{\pm2.1}$ & $57.2_{\pm0.1}$ & $20.0_{\pm4.7}$ & $64.4_{\pm0.8}$ & $24.6_{\pm2.4}$ & $52.9_{\pm0.2}$  \\
        Vaccine & $5.4_{\pm0.7}$ & $23.2_{\pm1.0}$ & $21.0_{\pm2.6}$ & $52.7_{\pm0.5}$ & $12.1_{\pm1.6}$ & $63.3_{\pm0.3}$ & $12.8_{\pm1.6}$ & $46.4_{\pm0.4}$  \\
        Booster & $5.9_{\pm0.2}$ & $44.0_{\pm0.9}$ & $29.0_{\pm0.1}$ & $57.3_{\pm0.1}$ & $10.9_{\pm0.0}$ & $68.8_{\pm0.0}$ & $15.2_{\pm0.1}$ & $56.6_{\pm0.3}$  \\
        LDIFS & $4.1_{\pm0.7}$ & $16.7_{\pm0.8}$ & $2.9_{\pm0.2}$ & $36.2_{\pm0.8}$ & $9.3_{\pm1.6}$ & $63.9_{\pm0.9}$ & $5.4_{\pm0.4}$ & $38.9_{\pm0.3}$  \\
        SafeInstruct & $6.6_{\pm0.3}$ & $42.5_{\pm0.9}$ & $9.1_{\pm0.0}$ & $58.7_{\pm0.1}$ & $6.9_{\pm0.3}$ & $60.1_{\pm0.3}$ & $7.5_{\pm0.0}$ & $53.8_{\pm0.4}$  \\
        Lisa & $5.3_{\pm0.1}$ & $38.9_{\pm0.9}$ & $7.6_{\pm0.0}$ & $54.4_{\pm0.3}$ & $2.7_{\pm1.4}$ & $52.6_{\pm7.8}$ & $5.2_{\pm0.1}$ & $48.6_{\pm2.5}$  \\
        Antidote & $12.9_{\pm7.6}$ & $47.0_{\pm1.1}$ & $5.9_{\pm4.1}$ & $60.3_{\pm0.8}$ & $4.5_{\pm0.5}$ & $61.0_{\pm0.7}$ & $1.5_{\pm0.3}$ & $60.5_{\pm0.1}$  \\
        \midrule
        SFT* & $3.4_{\pm2.8}$ & $41.7_{\pm1.2}$ & $1.3_{\pm0.2}$ & $57.7_{\pm0.5}$ & $1.7_{\pm0.3}$ & $59.7_{\pm0.9}$ & $2.1_{\pm0.9}$ & $53.0_{\pm0.6}$  \\
        Repnoise* & $6.2_{\pm6.1}$ & $37.2_{\pm0.5}$ & $1.7_{\pm0.3}$ & $57.1_{\pm1.1}$ & $2.3_{\pm0.1}$ & $63.4_{\pm0.9}$ & $3.4_{\pm2.0}$ & $52.6_{\pm0.7}$  \\
        Vaccine* & $1.4_{\pm0.4}$ & $22.4_{\pm0.5}$ & $1.1_{\pm0.1}$ & $53.0_{\pm0.4}$ & $1.7_{\pm0.2}$ & $62.6_{\pm0.6}$ & $1.4_{\pm0.1}$ & $46.0_{\pm0.2}$  \\
        Booster* & $1.9_{\pm1.1}$ & $43.6_{\pm0.1}$ & $1.1_{\pm0.1}$ & $57.1_{\pm0.7}$ & $2.1_{\pm0.3}$ & $69.0_{\pm0.1}$ & $1.7_{\pm0.4}$ & $56.6_{\pm0.2}$  \\
        LDIFS* & $1.6_{\pm0.9}$ & $16.9_{\pm0.4}$ & $1.1_{\pm0.1}$ & $36.9_{\pm1.1}$ & $1.6_{\pm0.3}$ & $62.7_{\pm1.3}$ & $1.4_{\pm0.4}$ & $38.8_{\pm0.5}$  \\
        SafeInstruct* & $1.1_{\pm0.1}$ & $43.1_{\pm0.5}$ & $1.2_{\pm0.1}$ & $58.8_{\pm0.2}$ & $1.6_{\pm0.3}$ & $59.3_{\pm0.5}$ & $1.3_{\pm0.2}$ & $53.7_{\pm0.4}$  \\
        Lisa* & $1.5_{\pm0.4}$ & $37.5_{\pm1.5}$ & $\textbf{1.0}_{\pm0.1}$ & $54.8_{\pm0.6}$ & $1.3_{\pm0.2}$ & $50.3_{\pm1.0}$ & $1.3_{\pm0.1}$ & $47.6_{\pm0.0}$  \\
        Antidote* & $1.3_{\pm0.1}$ & $46.1_{\pm0.6}$ & $\textbf{1.0}_{\pm0.1}$ & $60.6_{\pm0.7}$ & $3.3_{\pm2.0}$ & $61.2_{\pm0.9}$ & $1.9_{\pm0.7}$ & $56.0_{\pm0.1}$  \\
        % \cmidrule{1-9}
        % Deep Ignorance & $14.4_{\pm0.6}$ & $42.4_{\pm0.2}$ & $21.3_{\pm1.2}$ & $58.3_{\pm0.8}$ & $4.9_{\pm0.2}$ & $55.9_{\pm0.7}$ & $13.5_{\pm0.5}$ & $52.2_{\pm0.3}$  \\
        % Pharmacist*~\cite{liu2025pharmacist}) & 59.8 & 37.0 & 62.9 & 57.0 & 54.4 & 57.1 & 59.0 & 50.4  \\
        \cmidrule{1-9}
        Ours (Base) & $1.2_{\pm0.7}$ & $47.9_{\pm0.8}$ & $1.9_{\pm0.2}$ & $\textbf{63.8}_{\pm1.3}$ & $1.5_{\pm0.3}$ & $69.9_{\pm1.0}$ & $1.5_{\pm0.3}$ & $60.5_{\pm0.1}$  \\
        Ours (Ref-Teacher) & $\textbf{1.0}_{\pm0.5}$ & $\textbf{49.0}_{\pm0.5}$ & $1.2_{\pm0.2}$ & $62.4_{\pm1.1}$ & $\textbf{0.9}_{\pm0.3}$ & $\textbf{70.6}_{\pm0.8}$ & $\textbf{1.0}_{\pm0.3}$ & $\textbf{60.7}_{\pm0.2}$  \\
        \bottomrule
        \end{tabular}
        \vspace{-1mm}
        \caption{Performance across model architectures. Results on Llama3-8B, Gemma2-9B, and Qwen2-7B evaluate adaptability beyond a single backbone. Methods marked with * use LLaMAGuard3-8B-based data filtering.}
        \vspace{-2mm}
        \label{tab:architecture_filtered}
        % \end{adjustbox}
        % \vspace{-3mm}
\end{table*}

\section{Experiment}
\label{sec:metrics}
We evaluate the effectiveness of our paradigm across diverse safety and user-task settings by varying the harmful prompt ratio, user data size, harmless task dataset (GSM8K~\cite{cobbe2021trainingverifierssolvemath}, SST2~\cite{socher2013recursive}, AGNEWS~\cite{zhang2015character}, AlpacaEval~\cite{alpaca_eval}), and the base model (Llama3-8B~\cite{grattafiori2024llama}, Gemma2-9B~\cite{team2024gemma}, Qwen2-7B~\cite{yang2024qwen2}). Unless noted otherwise, we used Llama3-8B, $0.1$ poison ratio, $1,000$ user data, and GSM8K as harmless data.

\paragraph{Datasets.} For teacher preparation stage, we used $N=5,000$ harmful prompts paired with refusal responses from BeaverTails~\cite{ji2023beavertails}, and $N=5,000$ harmless prompts with helpful responses from Alpaca~\cite{alpaca}. For finetuning stage, user data was constructed by mixing harmful and harmless samples with a specific poison ratio. 
The alignment data size $N_\text{safety}$ was set equal to the user data size $N_\text{user}$. 
All harmful prompts were sourced from BeaverTails, but distinct subsets were used for the teacher preparation, finetuning, and evaluation to avoid data overlap.

\paragraph{Metrics.} 
We evaluate safety and utility using two metrics: Harmful Score (HS) and Finetuning Accuracy (FA), following prior work~\cite{huang2025antidote,huang2025booster}. HS is the proportion of harmful responses among $1,000$ outputs generated from BeaverTails test set, classified by Beaver-Dam-7B~\cite{ji2023beavertails}. FA is measured by downstream benchmarks for GSM8K, SST2, AGNEWS, and AlpacaEval. 

\paragraph{Baselines.}
% We compare our framework against alignment ({RepNoise}~\cite{rosati2024representation}, {Vaccine}~\cite{huang2024vaccine}, {Booster}~\cite{huang2025booster}), finetuning ({LDIFS}~\cite{mukhoti2025fine}, {SafeInstruct}~\cite{bianchi2024safety} {Lisa}~\cite{huang2024lisa}), post-finetuning-stage ({Antidote}~\cite{huang2025antidote}) solutions. We also include data-filtering baseline \textbf{Deep Ignorance}~\cite{o2025deep}.
% For fair comparison, we apply data filtering to baselines that do not originally include it, using \textbf{LlamaGuard3-8B} finetuned on the BeaverTails and Alpaca. Details for baselines and filtering are provided in Appendix~\ref{sec:baseline details}.
We compare our framework with representative alignment stage (RepNoise~\cite{rosati2024representation}, Vaccine~\cite{huang2024vaccine}, Booster~\cite{huang2025booster}), finetuning stage (LDIFS~\cite{mukhoti2025fine}, SafeInstruct~\cite{bianchi2024safety}, Lisa~\cite{huang2024lisa}), and a post-finetuning stage  (Antidote~\cite{huang2025antidote}) solutions. For fair comparison, we apply data filtering to the baselines using LLaMAGuard3-8B~\cite{grattafiori2024llama} finetuned on BeaverTails and Alpaca, which achieves filtering accuracy comparable to the Ref-Teacher. We report two variants of our paradigm, \textbf{Ours (Base)} and \textbf{Ours (Ref-Teacher)}, without and with Ref-Teacher guidance, respectively.

% Further details are provided in Supplementary Material.

\subsection{Experiment Results}

% \begin{table*}[!t]
%     \centering
%     % \begin{adjustbox}{max width=\textwidth}
%     \scriptsize
%     \begin{tabular}{l|ccc|cc}
%     \toprule
%     Harmful Datasets & BeaverTails & JailbreakBench & Toxic-chat & GCG & AutoDAN-turbo \\
%     \midrule
%     Harmless Datasets & \multicolumn{5}{c}{JailbreakBench} \\
%     \midrule
%     Linear Classifier & 83.5 & 69.8 & 75.7 & 52.4 & 48.4 \\
%     LLaMAGuard3-8B & 64.1 & \textbf{88.7} & 57.0 & 89.7 & 9.3 \\
%     OpenAI Moderation & 67.8 & 74.7 & 44.4 & 81.0 & 52.2 \\
%     % Deep Ignorance & & & & \\
%     Ref-Teacher ($\tau=0$) & \textbf{93.4} & 79.8 & \textbf{87.0} & \textbf{92.9} & \textbf{82.1}  \\
%     \bottomrule
%     \end{tabular}
%     \vspace{-3mm}
%     \caption{F1 Scores (\%) of Ref-Teacher and guardrail models across various jailbreaking attacks.}
%     \label{tab:classification_generalization}
%     \vspace{-2mm}
% \end{table*}

\paragraph{Robustness under Varying Harmful Ratio.} 
We evaluate our framework using HS and FA under varying ratios of harmful prompts $p$, ranging from fully clean user data ($p=0$) to moderately contaminated data ($p=0.5$). As shown in Table \ref{tab:harmful_ratio_filtered}, our methods consistently achieve the lowest HS and the highest FA across all values of $p$, outperforming all baselines. Although most methods report similar HS due to the application of data filtering, our approach attains noticeably higher finetuning accuracy. This stems from a structural difference in our FaaS framework: rather than finetuning a safety-aligned model whose weights have already been shifted toward safety, we finetune the base model under the guidance of a safety-aligned teacher. This preserves a more adaptable initialization while enforcing safety during finetuning, leading to a better safety-utility trade-off.

\paragraph{Scalability with Varying Amounts of User Data.}
We evaluate the scalability of our framework by measuring HS and FA as the number of user samples increases from $1{,}000$ to $2{,}500$. For a fixed poison ratio, most baselines maintain low HS as the number of harmful prompts increases, while FA improves with more user data. As shown in Table~\ref{tab:num_samples_filtered}, our Ref-Teacher-guided finetuning strategy consistently achieves the best performance across most settings.

\begin{table}[t]
    \centering
    % \begin{adjustbox}{max width=\textwidth}
    \scriptsize
    \begin{tabular}{l|cc|c}
    \toprule
    Datasets & Harmful & Harmless & Total \\
    \midrule
    GSM8K & 100.00 & 97.70 & 97.93  \\
    SST2 & 99.91 & 95.30 & 95.76  \\
    AGNEWS & 99.91 & 99.86 & 99.87 \\
    AlpacaEval & 99.90 & 77.04 & 79.33  \\
    \bottomrule
    \end{tabular}
    \vspace{-2mm}
    \caption{Prompt classification accuracy (\%) of Ref-Teacher during finetuning. We report harmful, harmless, and total accuracy across different user-task datasets.}
    \label{tab:classification_FT}
    \vspace{-2mm}
\end{table}

\begin{table}[!t]
    \centering
        \begin{adjustbox}{max width=\linewidth}
        \scriptsize
        \begin{tabular}{l|ccc|cc}
        \toprule
        Harmful Datasets & BeaverTails & JailbreakBench & Toxic-chat & GCG & AutoDAN-turbo \\
        \midrule
        Harmless Datasets & \multicolumn{5}{c}{JailbreakBench} \\
        \midrule
        Linear Classifier & 83.5 & 69.8 & 75.7 & 52.4 & 48.4 \\
        LLaMAGuard3-8B & 64.1 & \textbf{88.7} & 57.0 & 89.7 & 9.3 \\
        OpenAI Moderation & 67.8 & 74.7 & 44.4 & 81.0 & 52.2 \\
        Ref-Teacher ($\tau=0$) & \textbf{93.4} & 79.8 & \textbf{87.0} & \textbf{92.9} & \textbf{82.1}  \\
        \bottomrule
        \end{tabular}
        \end{adjustbox}
        \vspace{-2mm}
        \caption{F1 scores (\%) for cross-dataset prompt classification. Ref-Teacher achieves higher F1 scores than safety guardrail baselines across diverse harmful datasets.}
        \label{tab:classification_generalization}
        \vspace{-2mm}
\end{table}

\begin{table}[t]
    \centering
        \begin{adjustbox}{max width=\linewidth}
        \centering
        \vspace{5mm}
        % \begin{adjustbox}{max width=\textwidth}
        \scriptsize
        \begin{tabular}{cc|cc}
        \toprule
        AD & Filtering & HS $\downarrow$ & FA $\uparrow$ \\
        % \multirow{2}{*}{AD} & \multirow{2}{*}{Filtering} & \multicolumn{2}{c}{p=0.1} \\
        % \cmidrule{3-4}
        % & & HS $\downarrow$ & FA $\uparrow$ \\
        \midrule
        X & X & 2.0 & 47.9  \\
        O & X & 2.2 & 46.2 \\ % alpha 1
        % O & X & 10.2 & 46.7 \\ % alpha 0.1
        X & O & 0.6 & 46.5  \\
        O & O & 0.5 & 49.0  \\
        \bottomrule
        \end{tabular}
        \end{adjustbox}
        \vspace{-2mm}
        \caption{Ablation study on safety and utility.}
        \label{tab:ablation1}
        \vspace{-2mm}
\end{table}

\begin{table}[!t]
    \centering
        \begin{adjustbox}{max width=\linewidth}
        \centering
        % \begin{adjustbox}{max width=\textwidth}
        \scriptsize
        \begin{tabular}{cc|cc|cc|cc|cc}
        \toprule
        \multirow{2}{*}{AD} & \multirow{2}{*}{Filtering} & \multicolumn{2}{c}{$p=0$} & \multicolumn{2}{c}{p=0.1} & \multicolumn{2}{c}{p=0.3} & \multicolumn{2}{c}{p=0.5} \\
        \cmidrule{3-10}
          & & Freq (\%) & Avg &  Freq (\%) &  Avg &  Freq (\%) &  Avg  &  Freq (\%) &  Avg  \\
        \midrule
         X & X & 35.09 & 0.110 & 36.80 & 0.099 & 40.80 & 0.073 & 46.03 & 0.039 \\ 
         O & X  & 32.26 & 0.131 & 34.02 & 0.117 & 37.78 & 0.090 & 42.55 & 0.055 \\ 
         X & O  & 36.11 & 0.102 & 36.51 & 0.097 & 37.80 & 0.087 & 39.91 & 0.073 \\ 
         O & O  & 30.02 & 0.140 & 29.60 & 0.143 & 28.93 & 0.145 & 28.29 & 0.149 \\ 
        \bottomrule
        \end{tabular}
        \end{adjustbox}
        \vspace{-2mm}
        \caption{Ablation study on gradient conflicts.}
        \label{tab:gradient conflict ablation}
        \vspace{-3mm}
\end{table}

\paragraph{Generalization across Diverse User Datasets.}  
To evaluate generalization across datasets, we replace the harmless portion of user data from GSM8K with SST2, AGNEWS, and AlpacaEval samples, and measure HS and FA. As shown in Table~\ref{tab:dataset_filtered}, our approach consistently yields low HS and high FA across all datasets.  
These results demonstrate the generalization of our framework, preserving both safety alignment and task performance across diverse downstream tasks.

% \paragraph{Adaptability across Model Architectures.}  
% We assess adaptability across model architectures by training the Ref-Teacher on Gemma2-9B and Qwen2-7B, and finetuning the corresponding base models on safety alignment and user data.  
% For each architecture, we select the optimal layer for extracting refusal features (Details are in Appendix~\ref{sec: optimal layer}).  
% Table~\ref{tab:architecture_filtered} shows that our method consistently reduces HS while improving FA across model architectures, demonstrating that our approach is not restricted to a single architecture.  

\paragraph{Adaptability across Models.}
We evaluate adaptability on Gemma2-9B and Qwen2-7B by training a Ref-Teacher and finetuning the corresponding base model with safety-alignment and user data. For each architecture, we select the layer used to extract the refusal feature. As shown in Table~\ref{tab:architecture_filtered}, our method achieves low HS and improves FA across architectures, showing robustness beyond a single backbone.

\subsection{Analysis}

\paragraph{Classification Performance of Ref-Teacher.}  
We evaluate Ref-Teacher's prompt classification ability during finetuning on GSM8K, SST2, AGNEWS, and AlpacaEval, where it achieves near-perfect accuracy on harmful prompts and consistently high accuracy on harmless ones (Table~\ref{tab:classification_FT}).
For generalization, we test on JailbreakBench~\cite{chao2024jailbreakbench} harmless prompts combined with harmful prompts from BeaverTails, JailbreakBench, Toxic-chat~\cite{lin2023toxicchat}, GCG~\cite{zou2023universal}, and AutoDAN-turbo~\cite{liu2024autodan}.  
We compare Ref-Teacher with LLaMAGuard3-8B\footnote{Table~\ref{tab:classification_FT} reports \textbf{zero-shot} LLaMAGuard3-8B for reference only, not the filtering model used in the main experiments. The filtering-model accuracy is reported in Supplementary Material.}, OpenAI Moderation\footnote{https://platform.openai.com/docs/guides/moderation}, and a linear classifier trained on Llama3-8B features using BeaverTails and Alpaca.
As shown in Table~\ref{tab:classification_generalization}, Ref-Teacher consistently achieves high F1 scores across diverse attacks by leveraging its refusal feature, allowing stronger generalization to unseen harmful datasets than safeguard models that rely on labeled data.
% These results highlight the accuracy and generalization of Ref-Teacher-guided classification for reliable data filtering.

% \paragraph{Ablation Study on Safety and Task Performance.}  
% We assess the impact of alignment distillation (AD) and data filtering (Filtering) on safety and task performance by removing each component.  
% As shown in Table~\ref{tab:ablation1}, AD alone improves neither safety nor finetuning accuracy, indicating that it cannot stabilize optimization when harmful prompts remain in user data. In contrast, Filtering alone reduces harmfulness but lowers finetuning accuracy due to reduced user data, which increases overfitting risk.  
% These results highlight their complementary roles: AD stabilizes optimization but requires filtered data, whereas Filtering reduces harmfulness but risks overfitting without distillation.  
% Their combination synergistically achieves strong task performance while preserving safety alignment.  

% \paragraph{Ablation Study on Gradient Conflicts.}  
% We evaluate the contributions of alignment distillation (AD) and data filtering (Filtering) on gradient conflicts by removing each component and varying the harmful ratio $p$.  
% Table~\ref{tab:gradient conflict ablation} shows that AD alone reduces conflicted parameters on clean data but loses effectiveness as $p$ increases, while Filtering alone stabilizes the frequency of conflicts but does not sufficiently mitigate it.  
% Consequently, AD and Filtering complement each other in our framework, mitigating gradient conflicts effectively under harmful finetuning attacks.  

\paragraph{Ablation Study.}
We analyze the contributions of alignment distillation (AD) and data filtering (Filtering) by removing each component and evaluating both safety, utility, and gradient conflicts under varying harmful ratios $p$. As shown in Tables~\ref{tab:ablation1} and~\ref{tab:gradient conflict ablation}, AD alone fails to improve safety or finetuning accuracy when harmful prompts remain, although it reduces gradient conflicts, with diminishing effectiveness as $p$ increases. In contrast, Filtering alone effectively reduces harmfulness and stabilizes conflict frequency, but lowers finetuning accuracy due to reduced user data and increased overfitting risk. These results highlight the complementary roles of the two components: AD stabilizes optimization but requires filtered data, while Filtering removes harmful prompts but risks overfitting. Their combination synergistically mitigates gradient conflicts and achieves strong safety and task performance under harmful finetuning attacks.

\paragraph{Local Loss Smoothness.} 
We analyze the effect of alignment distillation on local loss smoothness using the gradient Lipschitz estimate, $L \approx |\nabla \ell(\theta+\epsilon)-\nabla \ell(\theta)| / |\epsilon|$, averaged over model parameters. Alignment distillation reduces this estimate from 15.0 to 10.5, indicating a smoother loss surface. This aligns with the reduced gradient conflicts in Table~\ref{tab:gradient conflict ablation} and helps explain the improved training stability.

\paragraph{Supplementary Material.}
Supplementary Material provides additional analyses and experiments, verifying comparable filtering accuracy with finetuned LLaMAGuard3-8B and robustness across cross-dataset, jailbreak, pre-aligned LLM, safety-adjacent, and multi-evaluator settings.

\section{Conclusion}
% We identify a key limitation of standard two-stage Finetuning-as-a-Service (FaaS) pipeline, where providers first safety-align an LLM and then finetune the model on user data.
% We show that safety-aligned models offer weak initialization for downstream task learning, leading to suboptimal utility and degraded safety after user finetuning.  
% To overcome this, we propose a Refusal-Teacher-guided finetuning framework, which directly finetunes the unaligned base model on both user and safety-alignment data under the guidance of Ref-Teacher via alignment distillation and data filtering. 
% Extensive experiments demonstrate that our framework consistently achieves the lowest harmful scores and the highest finetuning accuracy across diverse settings, offering a practical defense against harmful finetuning attacks in FaaS.
We identify a key limitation of the standard two-stage FaaS pipeline, which uses safety-aligned weights as the initialization for user finetuning.
Our results show that this aligned-weight finetuning paradigm can limit downstream adaptation and lead to a suboptimal safety-utility trade-off.
To address this, we propose a safe FaaS finetuning paradigm that keeps the base model as the adaptable learner while enforcing safety through Ref-Teacher-guided alignment distillation and data filtering during finetuning.
Extensive experiments show that this paradigm reduces harmful outputs while improving user-task performance across diverse finetuning scenarios.
These results suggest that safe FaaS customization should move beyond safety-aligned-weight initialization toward teacher-guided safety enforcement during user adaptation.

% \section{Limitation}
% Our Ref-Teacher-guided finetuning framework relies on the Ref-Teacher model, which is trained using the refusal feature.  Consequently, its safety alignment could be compromised if adversarial attacks are designed to disrupt or manipulate the refusal feature. In such cases, the customized model finetuned under the guidance of a compromised Ref-Teacher may also inherit weakened safety alignment.  

% \section{Limitation}
% Our Ref-Teacher-guided finetuning framework relies on a Ref-Teacher model whose behavior is driven by the refusal feature. Consequently, if adversarial attacks explicitly manipulate or distort the refusal feature, the Ref-Teacher may provide unreliable filtering or alignment signals. In such cases, the base LLM finetuned under the guidance of a compromised Ref-Teacher may inherit weakened safety alignment. Addressing robustness against targeted refusal-feature manipulation remains an important direction for future work.

% \bibliography{aaai2027}


\end{document}